**Aya Kaysan Bahjat**
Informatics Institute for
Postgraduate Studies,
Baghdad, Iraq

# Automated document processing system for government agencies using DBNET++ and BART models

**Aya Kaysan Bahjat**

**DOI:** https://www.doi.org/10.33545/27075923.2025.v6.i2a.100

**Abstract**
An automatic document classification system is presented that detects textual content in images and classifies documents into four predefined categories (Invoice, Report, Letter, and Form). The system supports both offline images (e.g., files on flash drives, HDDs, microSD) and real-time capture via connected cameras, and is designed to mitigate practical challenges such as variable illumination, arbitrary orientation, curved or partially occluded text, low resolution, and distant text. The pipeline comprises four stages: image capture and preprocessing, text detection [1] using a DBNet++ (Differentiable Binarization Network Plus) detector, and text classification [2] using a BART (Bidirectional and Auto-Regressive Transformers) classifier, all integrated within a user interface implemented in Python with PyQt5. The achieved results by the system for text detection in images were good at about 92.88% through 10 hours on Total-Text dataset that involve high resolution images simulate a various and very difficult challenges. The results indicate the proposed approach is effective for practical, mixed-source document categorization in unconstrained imaging scenarios.

**Introduction**
In this article, we present automated document classification system that play a central role in modern information management by organizing, indexing, and routing large volumes of textual and scanned documents so that organizations can retrieve and act on information rapidly and reliably. Manual document classification often performed by human clerks or subject experts is time consuming, inconsistent, and prone to errors and bias, which degrades decision making and increases operational cost as document collections grow. To address these limitations, automated approaches leverage features extracted from document text, layout, and visual appearance to assign labels or categories with high throughput and reproducibility. Broadly, classification solutions range from traditional rule-based and classical machine learning pipelines (for example, feature engineering combined with Naive Bayes, logistic regression or Support Vector Machine (SVM), and etc.) to modern deep learning architectures that learn hierarchical and contextual representations directly from data (convolutional networks, recurrent models, and more recently transformer based encoders). Automated systems improve scalability, reduce human workload, and enable timely services such as intelligent search, automated routing, compliance checking, and analytics, while also reducing labeling inconsistencies and retrieval latency. Nevertheless, challenges remain heterogeneous document formats, imbalanced classes, noisy optical character recognition (OCR) output, and the need for labeled training data [3]. The remaining article summary illustrated as follow: in section (2) this field involve related works review, in section (3) a description of the architecture of the model used, in section (4) defines the experimental results of text detection that evaluated with Total-Text dataset (as shown in Fig. 1) [4], finally in section (5), and (6) we introduce a conclusion about the model with future directions for the beginners researchers.

**Related Work**
S. Long, *et al.*, in 2018 [5], Text Snake introduced a flexible scene text representation using sequences of overlapping disks along the text's central axis, predicting geometric attributes like radius and orientation through a fully convolutional network (FCN).

**Corresponding Author:**
**Aya Kaysan Bahjat**
Informatics Institute for
Postgraduate Studies,
Baghdad, Iraq

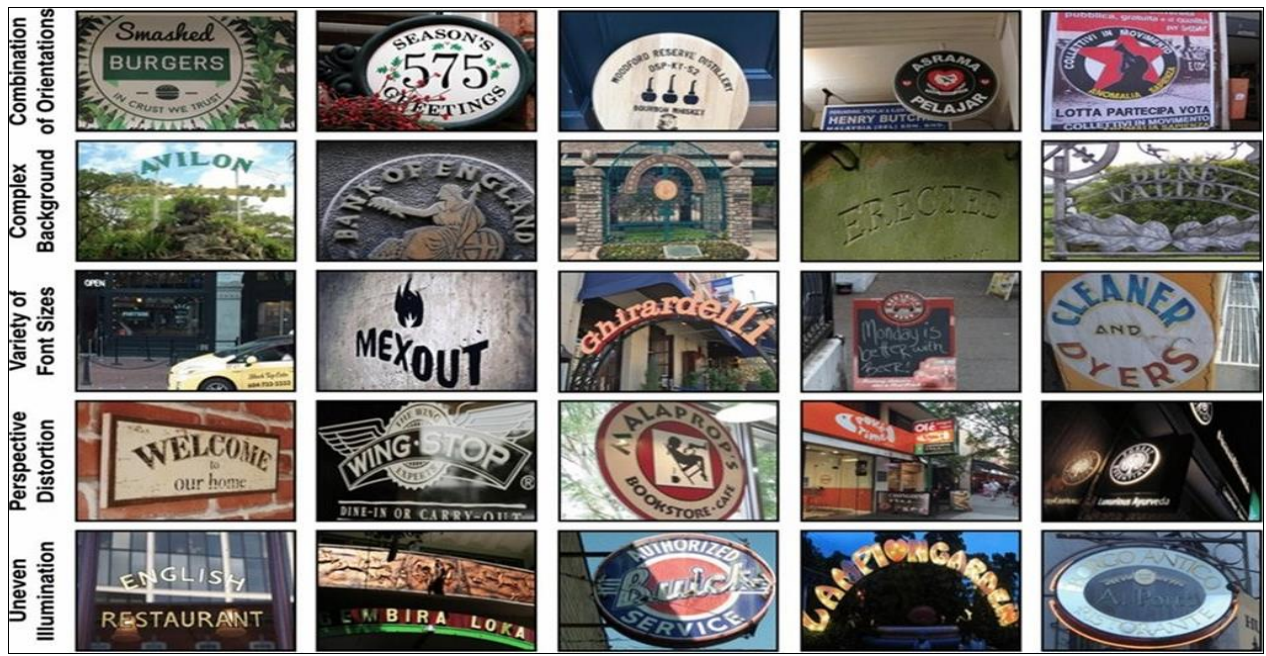

**Fig 1:** Over view of Total-Text Dataset [4].

The method excels in modeling curved text instances across multiple benchmarks, including MSRA-TD500, ICDAR2015, SCUT-CTW1500, and Total-Text. Impressively, Text Snake delivers more than a 40% F-measure improvement over the baseline on Total-Text highlighting its effectiveness in detecting free form text shapes, Y. Baek, *et al.*, in 2019 [6] they deploy character region awareness for text detection (CRAFT) that uses character center heatmaps and affinity mapping for the detection via segmentation.

Its F-measure on Total-Text is near 83.6%, effectively handling irregular text layouts. Meanwhile, subsequent studies investigate pre and post-processing enhancements such as blind deconvolution and automatic blur classification to further boost performance in natural scenes, J. Ye, *et al.*, in 2020 [7] The proposed system is using an instance segmentation based detector method that fuses character, word, and global level features. It reports a percentage F-measure of 87.1% on Total-Text dataset (ResNet-101, and single scale), S. Zhang, *et al.*, in 2021 [8] They used Adaptive Boundary Proposal Network for Arbitrary Shape Text Detection (ABPNet) frames arbitrary shaped text localization as an iterative boundary proposal and deformation process: a lightweight boundary proposal module produces coarse text contours and an adaptive deformation network (using Graph Convolutional Network (GCN)/Recurrent Neural Network (RNN) like components) refines those boundaries to fit text shapes precisely. The pipeline reduces reliance on heavy post-processing and improves boundary accuracy for highly curved words. On Total-Text test the (ABPNet) entry reports high precision and recall and an F-measure of 86.87%, showing that learned boundary deformation is an effective alternative to dense mask or polygon regression., F. Zhao, *et al.*, in 2022 [9] This method presents a backbone of ResNet-50 and an architecture combining global and word level cues for scene text detection. On the Total-Text, it achieves an impressive F-measure of 87.9%, outperforming prior state of the art detectors under comparable settings, Y. Su, *et al.*, in 2022 [10] Discrete Cosine Transform (DCT) encodes text instance masks into compact vectors using (DCT), paired with a lightweight anchor free detection frame-work. On Total-Text, it delivers an F-measure of percentage 84.9% at rate 15.1 FPS offering a commendable balance between accuracy and efficiency, Y. You, *et al.*, in 2023 [11] This method builds on a deformable Detection Transformer (DETR) architecture to directly predict control B-Spline points for each text contour, enabling smooth and flexible modeling of arbitrarily shaped words. It achieves with F-measure about 87.6% on Total-Text dataset, showcasing strong contour accuracy without pretraining data though at the cost of increased parameter complexity and computational demand, Z. Chen, *et al.*, in 2024 [12] Proposes a method named hyperbolic tangent binarization (HTB) plus multi scale channel attention and fused modules; achieves an accuracy ratio by adopts F-measure of 87.1 = 86.0% on the Total-Text dataset (ResNet-50).

We notice that most previous related works did not achieve high percentage of accuracy in the situation of the unconstraint conditions like illumination, orientation, curved text, partial occlusion, low resolution, and far away distance, therefore our system was evaluated on Total-Text dataset to simulate these conditions, where the main objective of this article is to get a high accuracy for text detection for both real-time and browsing modes.

**3. The Proposed Method:** This section illustrate the details of our project (as shown in Fig. 2),

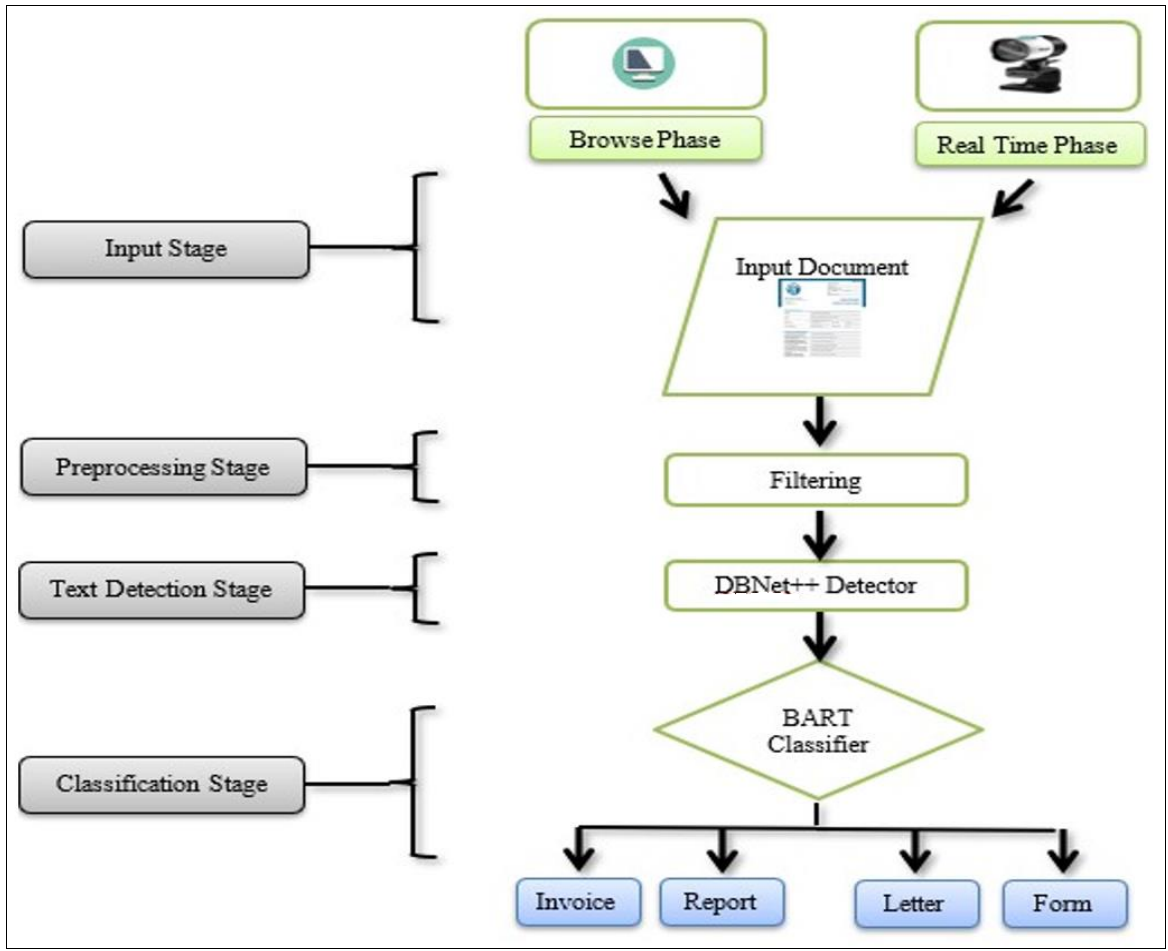

**Fig 2:** Automated Classification System Block-Diagram.

In our system, we utilize four pre labeled deferent classes (Invoice, Report, Letter, and Form), where the input documents firstly preprocessed by covert it from color to grayscale [13] and filtering with RealESRGAN [14] and CLAHE [15] filters, then the image texts are detected by Differentiable Binarization Network Plus (DBNet++) [16] model that utilize a ResNet-50 backbone and differentiable binarization (DB) module, which learns both a text probability map and threshold map in end to end fashion, and finally a model called Bidirectional and Autoregressive Transformer (BART) [17] that utilize encoder and decoder with Softmax function has been used for classifying the detected text. Once these models had been correctly implemented, the tasks like text classification and summarization can be easily implemented (as illustrated in Fig. 3). After that, the document will be classified into one of prior defined categories depending on the results of the classification.

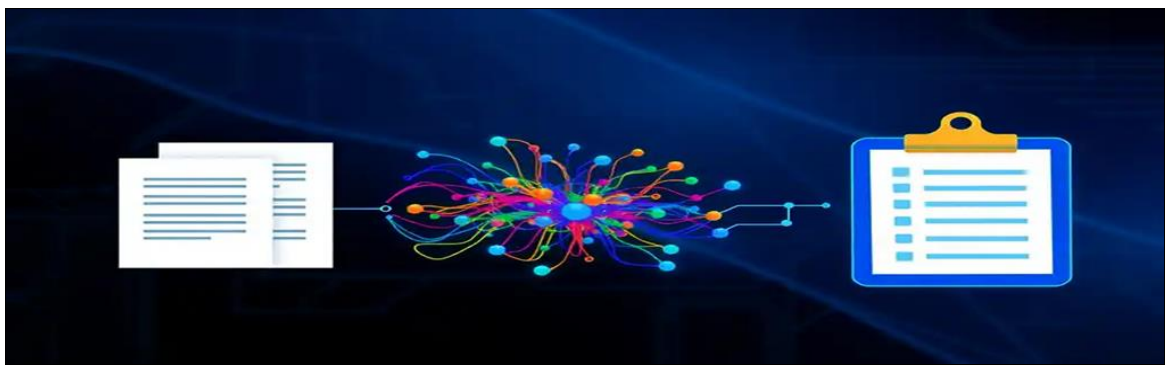

**Fig 3:** Over View of BART Rule [17].

## 3.1 Preprocessing Operations
We can be preprocessing the input document image by utilizing Real ESRGAN, and CLAHE filters as follow:

**1. Real ESRGAN**
In Fig. 4 after convert input document image into grayscale, Real ESRGAN [14] will increase its resolution.

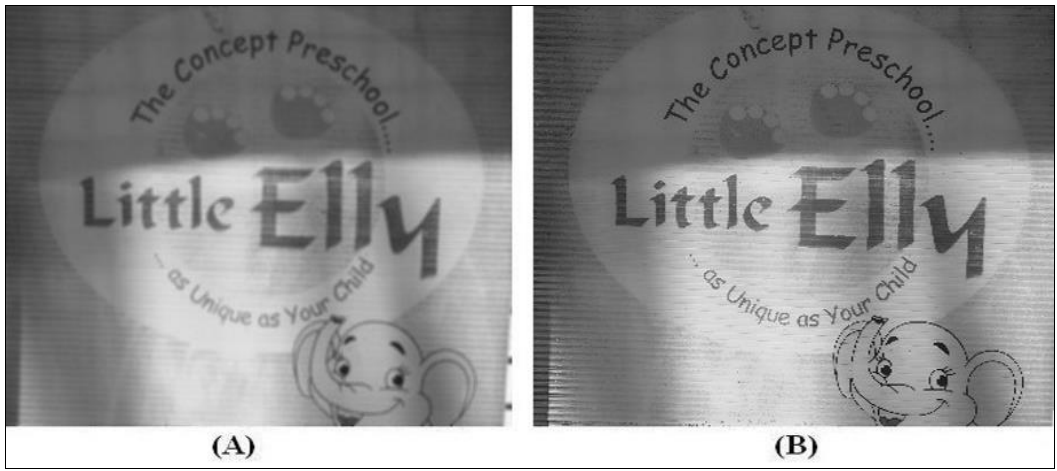

**Fig 4:** (A) Original Image (B) Applying Real ESRGAN Filter of the Image.

In Fig. 4 after convert input document image into grayscale, Real ESRGAN [14] will increase its resolution.

A (RealESRGAN_x2) super-resolution model is applied to increase image resolution by a factor of two so that small strokes, thin serifs and low-contrast character details become more legible for downstream OCR, as illustrated in the Proposed Method. Real-ESRGAN is a practical extension of ESRGAN that employs a Residual-in-Residual Dense Block (RRDB) generator trained with a realistic degradation pipeline so the model generalizes to common real-world artifacts (compression, sensor noise, and blur). The RRDB architecture and the degradation aware training make RealESRGAN particularly effective at restoring stroke continuity and contrast without producing strong hallucinated textures that would confuse (OCR) operation. In the proposed system we use the RealESRGAN_x2 variant to upscale text detection; chosen processing parameters are: scale Factor = 2, model = RealESRGAN_x2; real-time processing tiled inference with tile size = 64 and tile overlap = 16 to meet memory and latency constraints. These settings selected to balance restoration quality and computational cost, favoring conservative enhancement that increases character legibility while minimizing artificial detail.

**2. Contrast-Limited Adaptive Histogram Equalization (CLAHE)**

The next stage is applying CLAHE filter [15] that uses a clip factor of the image to prevent amplifying the noise in the uniform regions; in the histogram, if there is a peak, it is cut according to a pre-specified threshold. In this research, clip-limit = 8.0 with (8×8) tile size was adopted. This contrast enhancement intends to make text features more visible (as shown in Fig. 5), especially in the low illumination, and thus help the algorithm to better detect the features.

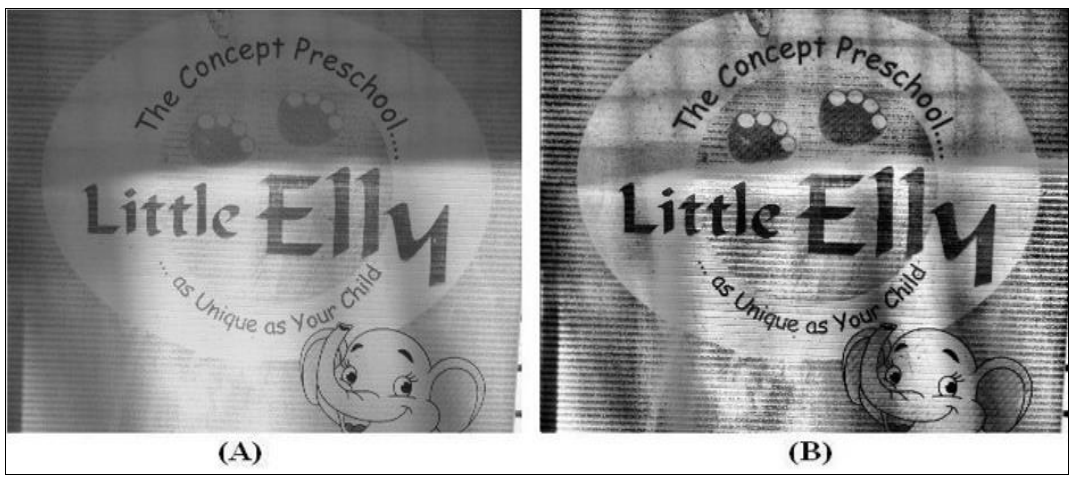

**Fig 5:** (A) Original Image (B) Applying CLAHE Filter of the Image.

**3.2 DBNet++ Detector**

Text detection in this project is performed using DBNet++ [16], is an advanced segmentation-based model designed for arbitrary-shaped text localization. DBNet++ first employs a (ResNet-50 backbone) with feature pyramid network (FPN) for extracting rich, multi scale representations from the input image, Then these feature maps are passed to the differentiable binarization module, which learns both a text probability map and threshold map in end to end fashion. Candidate text regions are identified by applying the learned soft threshold: pixels whose probability exceeds the corresponding threshold value are grouped into connected components, effectively delineating individual text instances rather than processing the entire image at once. In this project, a minimum text height of 5 pixels and a maximum of 1024 pixels were adopted to ensure robust detection across scales, and a default binarization threshold of 0.25 was used for initial candidate selection. DBNet++ was chosen for three main reasons: first, its differentiable binarization enables precise boundaries for curved and irregular text; second, its multi-scale feature aggregation allows reliable detection of both small and large text; and third, it delivers high detection accuracy with real-time inference speeds suitable for deployment. In this project, DBNet++ is implemented utilizing (PyTorch frame-work) in conjunction with the MMOCR toolbox. In (Fig. 6), an example of text detection using DBNet++ as illustrated.

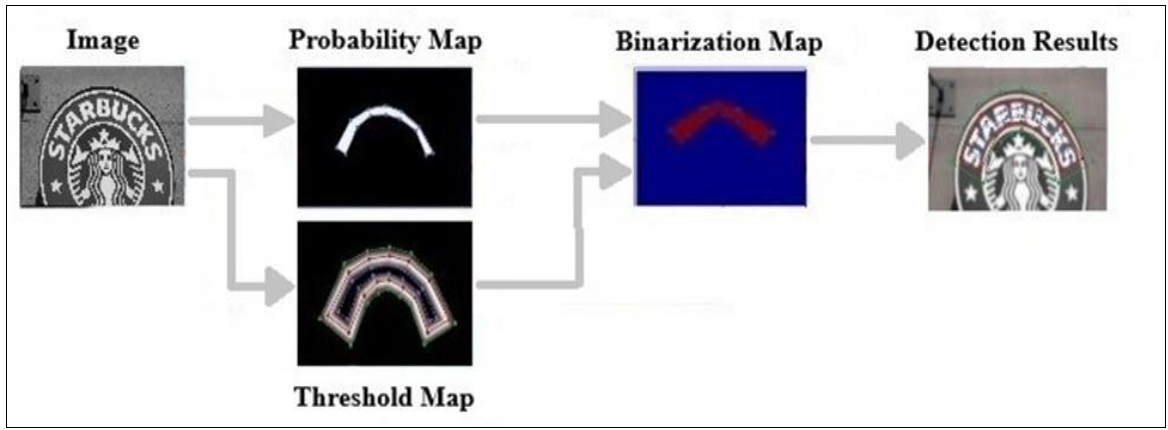

**Fig 6:** DBNet++ Detector.

### 3.3 BART Classifier
In the proposed system, the BART (Bidirectional and Auto-Regressive Transformers) [17] model is employed for text classification tasks by leveraging its encoder-decoder architecture. Specifically, the pre-trained model (facebook/bart-large-mnli) variant is used due to its adaptation for natural language inference (NLI), which enables zero-shot classification. During inference, the encoder processes the input sequence (e.g., a text document or sentence) into contextualized token embeddings using multiple self-attention layers. These embeddings are passed into the decoder, which has been fine-tuned for (NLI) by generating hypotheses such as "This text is about [label]". The decoder applies cross-attention between its layers and the encoder outputs, aligning generated tokens with source representations. A softmax layer at the output of the decoder computes the probability distribution over (NLI) labels entailment, contradiction, and neutral for each hypothesis. The classification decision is made by selecting the label with the highest entailment score. This mechanism allows BART to predict relevant class labels without explicit task-specific training, making it suitable for zero-shot classification scenarios in the proposed system (as shown in Fig. 7).

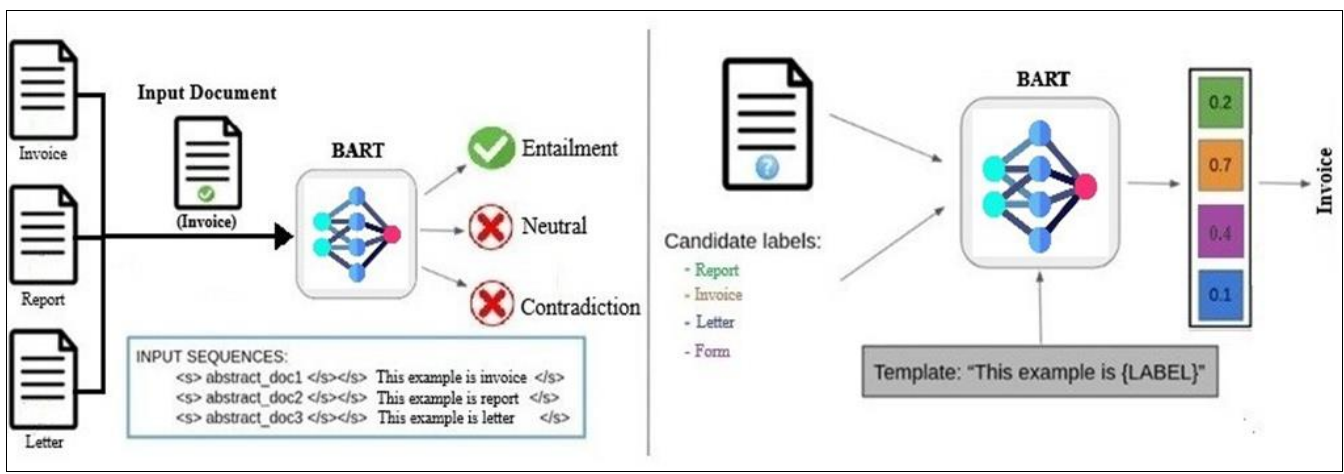

**Fig 7:** Bart Classifier.

### Experimental Result
In this proposed project, the pretrained models named DBNet++ was used for detect texts in images, and pretrained BART model (facebook/bart-large-mnli) for classifying the text and (facebook/bart-large-cnn) for summarize it, so there is no need to a training processing phase. In other side, the testing process phase took near by a ten hours to complete detecting all texts in the in all test images of the total-text dataset, and achieve 92.88% accuracy of text detection percentage (as shown in Fig. 8).

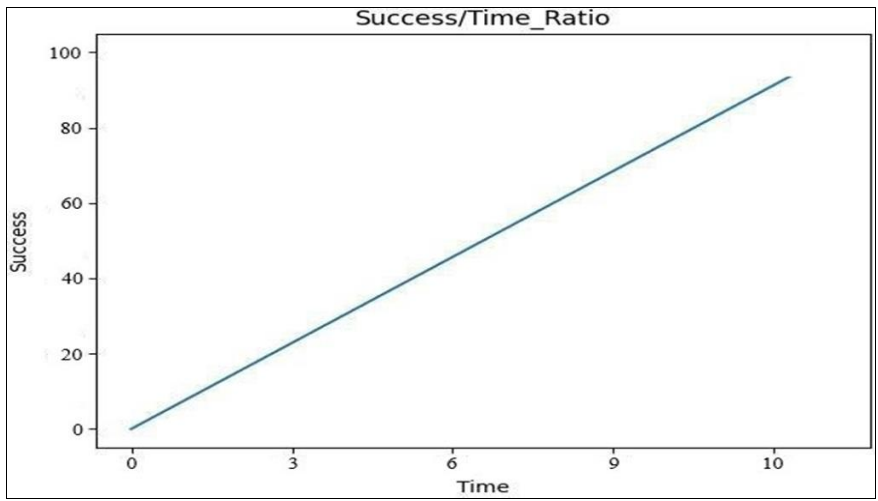

**Fig 8:** Time and Success Ratio evaluation.

We can utilize the computer internal web-cam or any other external web-cam. In our project, we utilized an external web-cam named Microsoft Life-cam Studio (shown in Fig. 9), for real-time experiments.

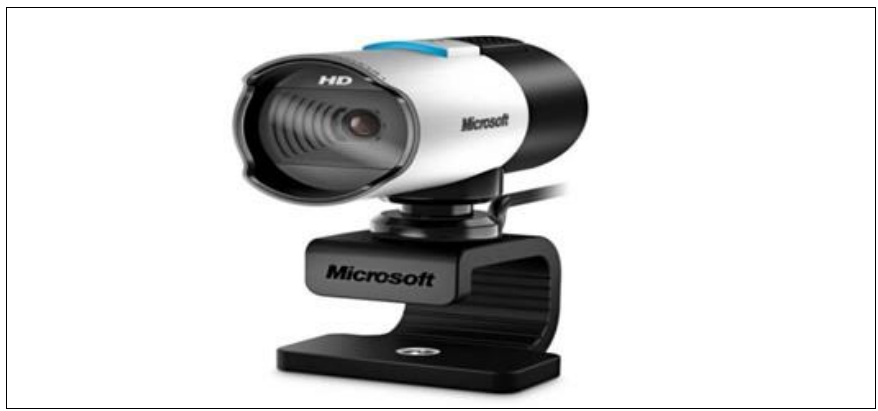

**Fig 9:** Microsoft Life-Cam Studio.

There is an practical example in (Fig. 10) for text classification in images browsing mode and another practical example in (Fig. 11) for text classification in real-time mode, respectively.

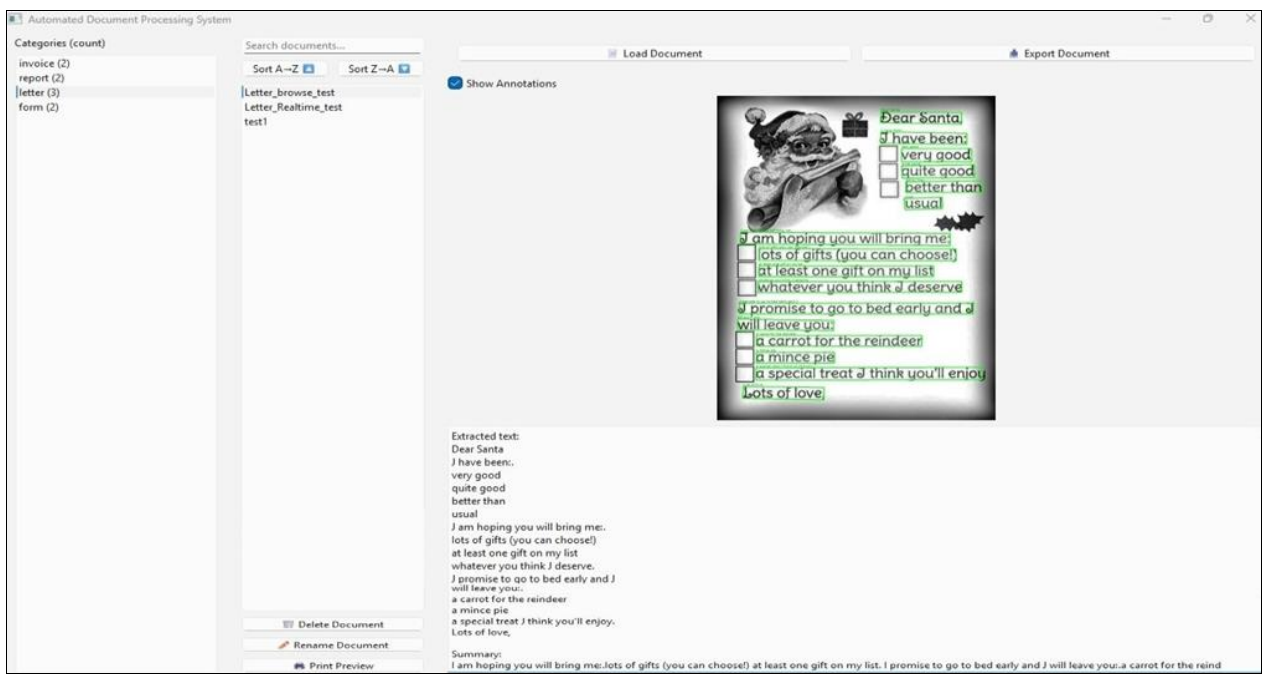

**Fig 10:** Classifying input document image in browsing mode example.

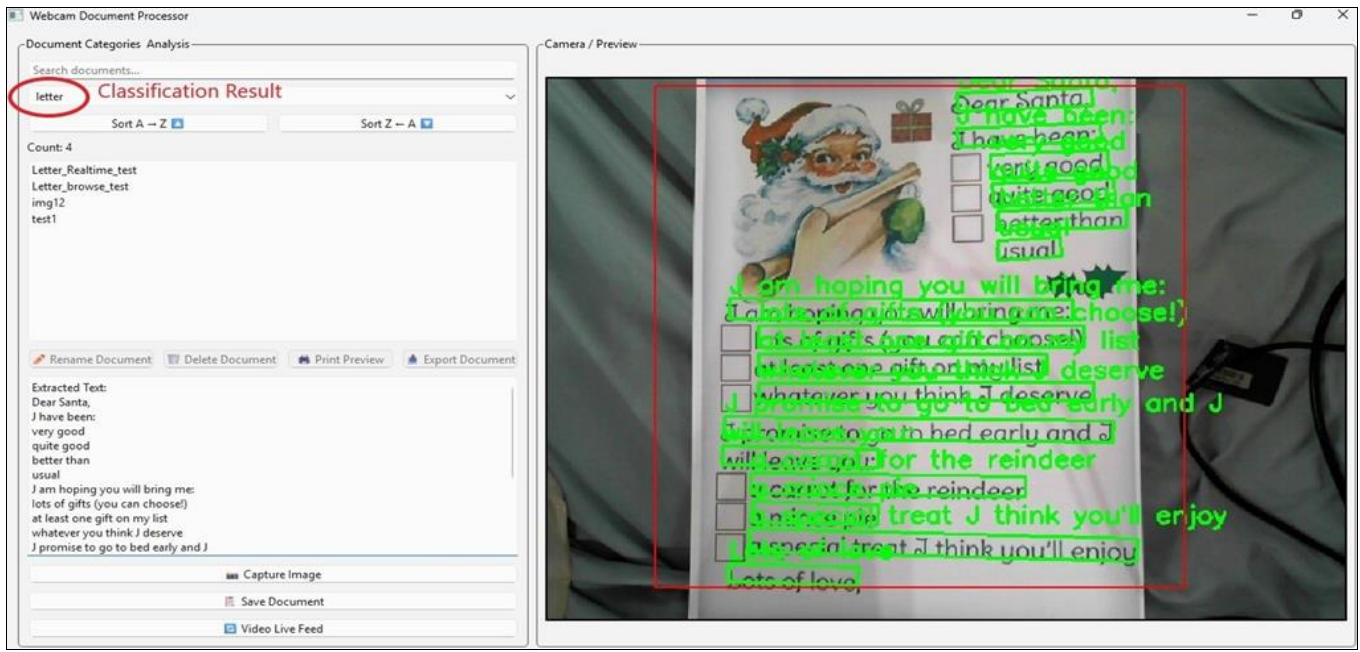

**Fig 11:** Classifying input document image in Real-time mode example.

## 5. Applications of Text Classification
Classifying documents helps organizations in many industries (as shown in Fig. 12). It improves their workflow, productivity, and compliance. There are some real-life examples, which show how the industry uses document classification are listed below [18]:

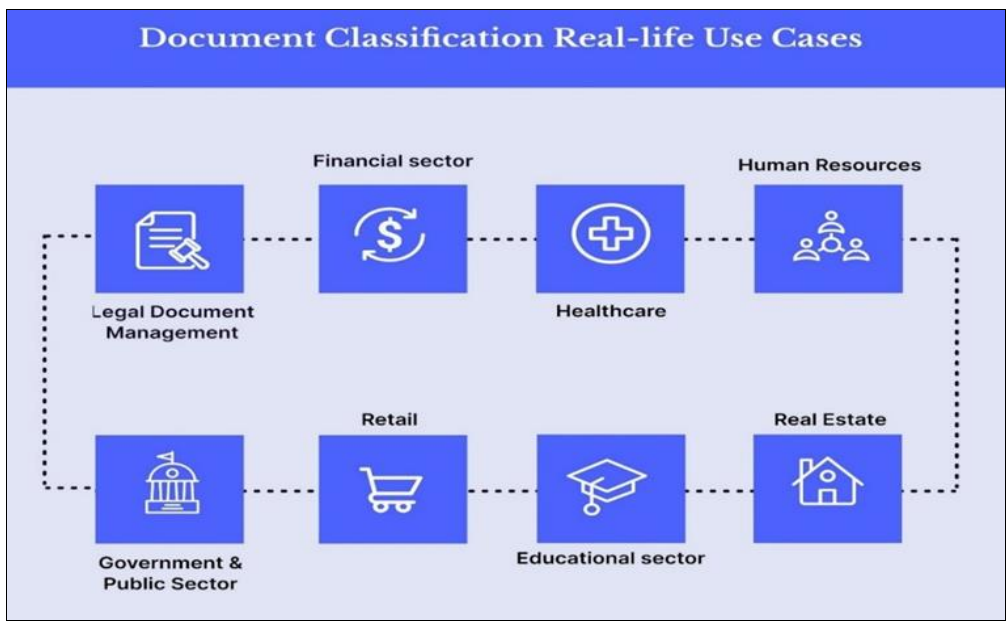

**Fig 12:** Documents Classification in Real-Life Use Cases [17].

1. **Legal:** classifies case files and filings, tags sensitive data, and improves indexing, electronic discovery, and compliance.
2. **Financial:** Sorts invoices, statements and loan docs to accelerate processing, validation and regulatory checks.
3. **Healthcare:** Groups medical records, claims and histories to speed claims, retrieval and clinical decision support.
4. **Human Resources (HR):** Presorts onboarding, payroll, training and evaluation records to streamline audits and (HR) workflows.
5. **Government and Public Sector:** Organizes legislative and regulatory documents to increase transparency, access, and compliance.
6. **Retail and E-Commerce:** Categorizes orders, queries and supplier documents to speed order processing, customer service and inventory control.
7. **Education:** Classify student's records, thesis and research to aid administration, planning, research discovery and accreditation.
8. **Real Estate:** Classifies property files, contracts and transaction records to reduce admin overhead and speed closing.

## 6. Conclusions
The method that suggested introduces a type of automatic text classification system that can be utilized in many government organizations environment. In the experimental result section, our system achieves new rate of accuracy about 92.88% in the Total-Text dataset for text detection in images. The proposed system proved that text detection and classification technologies used are suitable method in general institutions where it can be utilized in various implementations like human resources, financial, healthcare, education, and etc. The proposed system is suitable because it is a way for text detection, and classification with a very low consumption, especially it does not require external devices.

## 7. Future Works
This work can be expanded in various directions. Some ideas for future work are suggested below:

- **Domain adaptation and continual learning**
  Develop mechanisms to adapt the model to new document domains (legal, medical, financial, and etc.) with minimal labeled data: unsupervised domain adaptation, few-shot learning, and continual learning to avoid catastrophic forgetting when adding new classes.
- **Low resource and multilingual support**
  Building models that perform well for languages or scripts with limited labeled data (Arabic, Hindi, Chines, and etc.) using cross lingual transfer, synthetic data generation, and multilingual pretraining, with measuring per language performance and error modes.
- **Automatic taxonomy induction and label evolution.**
  Implement methods that automatically suggest new classes or merge or split existing classes based on clustering and user feedback, enabling the taxonomy to evolve with changing document collections.